# Printable Flexible Robots for Remote Learning

Savita V. Kendre,[1] Gus. T. Teran,[1] Lauryn Whiteside,[1] Tyler Looney,[1] Ryley Wheelock,[1] Surya Ghai,[1] and Markus P. Nemitz[1*]

The COVID-19 pandemic has revealed the importance of digital fabrication to enable online learning, which remains a challenge for robotics courses. We introduce a teaching methodology that allows students to participate remotely in a hands-on robotics course involving the design and fabrication of robots. Our methodology employs 3D printing techniques with flexible filaments to create innovative soft robots; robots are made from flexible, as opposed to rigid, materials. Students design flexible robotic components such as actuators, sensors, and controllers using CAD software, upload their designs to a remote 3D printing station, monitor the print with a web camera, and inspect the components with lab staff before being mailed for testing and assembly. At the end of the course, students will have iterated through several designs and created fluidically-driven soft robots. Our remote teaching methodology enables educators to utilize 3D printing resources to teach soft robotics and cultivate creativity among students to design novel and innovative robots. Our methodology seeks to democratize robotics engineering by decoupling hands-on learning experiences from expensive equipment in the learning environment.

**Introduction**

**A. Impact of remote learning in engineering courses**

Restriction of in-person classes during COVID-19 affected the hands-on learning experience of students from the fields of engineering and science [1]. Faculty in the engineering disciplines, such as robotics, have particularly struggled to teach students without in-person, hands-on laboratory sessions [2]. Robotics courses rely on the physical integration of systems which becomes difficult to accomplish when access to lab equipment is limited [3]. The ongoing COVID-19 pandemic has revealed the importance of a sustainable remote education system that allows the combination of theory and practice. Digital fabrication techniques are becoming a valuable tool to provide hands-on experience for students during online learning.

**B. Importance of design and fabrication in robotics engineering**

Robotics is a multi-disciplinary field that includes an understanding of mechanics, electronics, and computer science [4]. Any conventional, rigid robotic system is made from electro-mechanical and electronic parts such as motors, power electronics, and controllers. Conventional robots are typically purchased off-the-shelf and are being programmed to study robot behaviors or specific tasks (e.g., welding car joints in the auto industry). When roboticists develop conventional robots, they design, fabricate, and assemble robotic links and joints, and integrate sensors and electronics throughout the robot body.

---

[1]Department of Robotics Engineering, Worcester Polytechnic Institute, 100 Institute Road, Worcester, MA 01609, USA.
*Corresponding author. Email: mnemitz@wpi.edu

For educational purposes, there are robotic kits available that allow students to assemble their own robots [5]. These robotic kits can be challenging to use for remote education as the failure of a single component can affect the functionality of an entire robot. The ongoing shortage of silicon chips may delay the delivery of electronic components. In general, limitations on rapid shipment of replacement parts and restrictions to lab equipment such as oscilloscopes can further frustrate and hinder students from making progress. In contrast to conventional robots, soft robots, made from flexible and stretchable materials, do not require as many electronic or electro-mechanical components as rigid robots. Soft robots can be made from a single material, allowing sensors, actuators, and controllers to operate using pressurized air instead of electricity.

**C. Replica molding of pneumatically driven soft robots**

Soft robots have shown a wide variety of advantages compared to rigid robots including low-cost, safety, compatibility in interaction with humans and animals, and resistance to impact or corrosive chemicals, among others [6]. Soft robots have been actuated in numerous ways including magnetics, electro-statics, hydraulics, and pneumatics. Pneumatics has been the most commonly used actuation strategy; cavities made from stretchable materials expand upon pressurization, which can be used for locomotion, gripping of delicate objects, or shape-changing, among others [7].

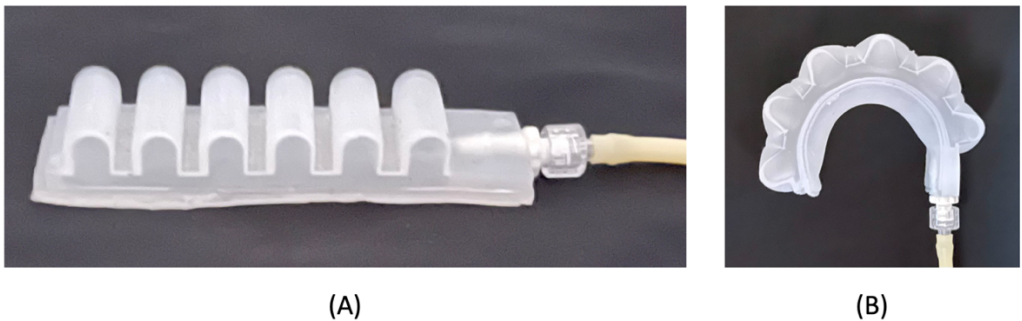

**Figure 1: Single-channel pneumatic network (PneuNet) fabricated using replica-molding.**
(A) Unactuated state of PneuNet. (B) Actuated state of PneuNet at 10 kPa air pressure.

One example of a pneumatic soft robotic actuator is the pneumatic network (PneuNet) **(Figure 1)**. PneuNets consist of networks of inflatable chambers that are interconnected with fluidic channels [8]. The chambers are similar to inflatable balloons, except that they can withstand large strains (> 900%) without plastically deforming. When positive pressure is applied, the chamber inflates and deforms based on the patterning profile **(Figure 1B)**. Hence, the design of the pneumatic network determines the directional expansion [9].

PneuNets are typically fabricated using replica molding (**Figure 2A**). First, the design of a PneuNet is created using CAD software (e.g., Autodesk Inventor). Second, the negative image of the PneuNet (i.e., mold) is fabricated using a 3D printer or CNC machine. Third, a commercially available additive curing elastomer is mixed, degassed in a vacuum chamber, and poured into the mold. The degassing process of the elastomer avoids air bubbles from being cast into the mold. Fourth, the elastomer needs to cure, whereas the curing time depends on the type of elastomer used and curing temperature. Last, the top and bottom casted parts are assembled using additional additive curing elastomer or epoxy.

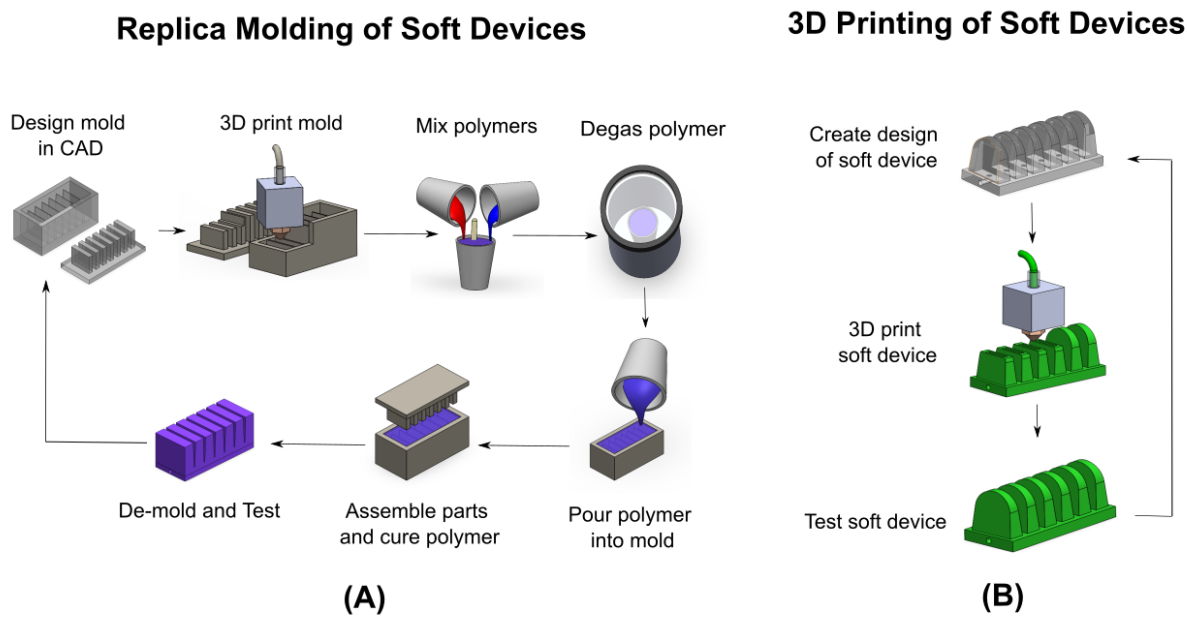

**Figure 2: Comparison of fabrication techniques for fluidically-driven soft devices** (A) Fabrication steps for replica molding to create a soft device. (B) Fabrication steps for 3D printing of a soft device.

The exploration of designs and characterization of components using replica molding is laborious and time-consuming due to the multi-step fabrication process. Most importantly, replica molding requires specialized equipment such as a vacuum chamber and a workspace for casting elastomers including personal protective equipment such as lab coats and gloves, making it a challenging fabrication technique to adopt in a remote classroom. Recent work on 3D printing soft materials, however, shows promise towards simplifying the fabrication process of soft devices. By printing multiple flexible materials with a single printer, it is possible to create complex designs using a single-step fabrication process.

**D.  3D printing of pneumatically driven soft robots.**

3D printing, a form of additive manufacturing, is a process of rapid deposition of materials to create an object by layer-wise controlled deposition through software. There are different types of 3D printers and printing protocols, which allow increasingly complex designs. Wallin et al. discuss common 3D printing techniques for soft robots including fused deposition modeling (FDM), stereolithography (SLA), direct ink writing (DIW), and selective laser sintering (SLS), among others [10]. These techniques use filaments, inks, powders, or resins as printing materials.

FDM printing is the most common 3D printing fabrication technique; it uses the melting of a filament to construct an object in successive layers [10]. For FDM printing soft devices, first, the design of components is created in CAD software, then it is printed using flexible filaments, and last, the component is tested (**Figure 2B**). In comparison to replica molding, 3D printed flexible filaments solidify at the time of printing; 3D prints can be used directly after fabrication. While replica molding is a multi-step fabrication process (**Figure 2A**), 3D printing allows for a single-step fabrication of soft robots (**Figure 2B**).

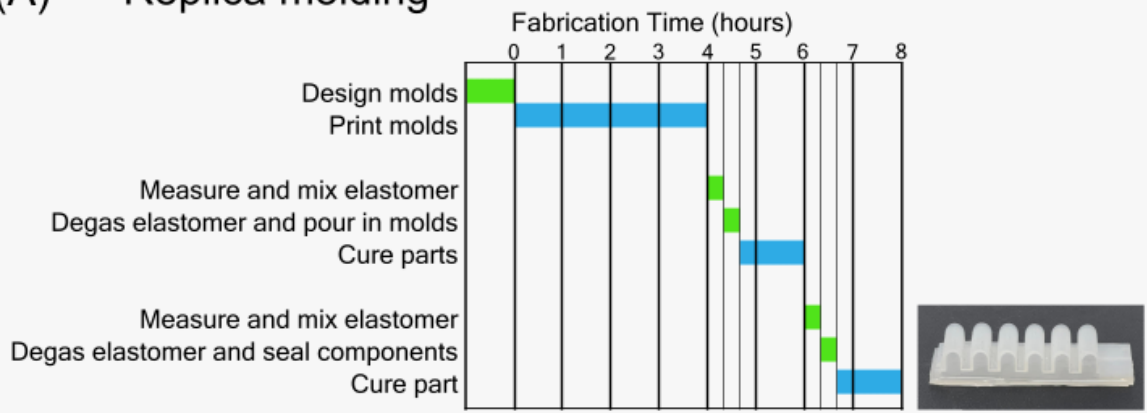
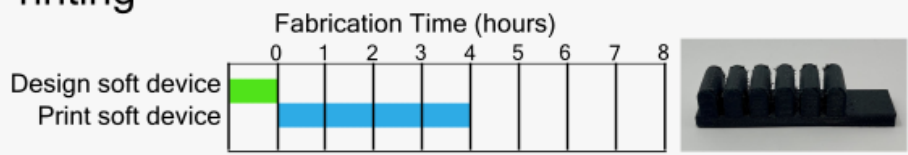

**Figure 3: Comparison of fabrication times of a PneuNet with replica molding and FDM printing.** The graphs detail the time distribution of fabrication steps in both, replica molding and FDM printing. (A) Replica molding a PneuNet takes approximately 8 hours, assuming perfect timing between fabrication steps. (B) FDM printing a PneuNet takes approximately 4 hours. We excluded the time for *designing* a PneuNet from the overall fabrication time. Green bars indicate manual fabrication steps executed by a person; blue bars indicate tasks that can be left unattended.

For example, YAP. et al. demonstrated the fabrication of a single and dual-channel pneumatically activated linear actuator on a FDM printer with flexible filament [11]. Wallin et al. showcased the use of a multi-material 3D printer to integrate soft actuators and sensors [12]. The fabrication of a single-channel PneuNet with replica molding takes ~8 hours, whereas it can be 3D printed in ~4 hours (**Figure 3**).

The increasing use of FDM printing has led to a wide range of printers that fit diverse needs, many supported filaments, decreasing costs, and comprehensive online support. FDM printers are low-cost such as the Prusa Mini+ ($399), Prusa MK3S ($899), or Creality Ender 3 Pro ($189), and have been tested with flexible filaments to create soft actuators, sensors, and controllers. FDM printers can be easily set up to a remotely accessible printing service including webcams and open-source software (e.g., OctoPrint), therewith allowing elastomeric robots and their components to be printed remotely. Our methodology uses FDM printing among other 3D printing techniques due to its low-cost, easy adoption, commercial availability, a wide range of materials, and multi-material and remote printing capabilities.

**Methodology**

We developed a remote learning methodology using printable, flexible robots to cultivate a hands-on learning experience for robotics engineering students (**Figure 4**). We divide our methodology into four sections (i) design of printable robots, (ii) web-based interface for remote printing and monitoring, (iii) inspection and delivery by teaching staff, and (iv) assembly and testing at home.

A.  **Design of printable robots**

Students design pneumatically driven soft robots through iterative design. They create designs for their robots comprised of fluidic sensors, actuators, control logic, and channels, using existing sample designs (e.g., 3D printed single-channel actuators [11]) from academic literature. Students will learn to design CAD models of their components with consideration to FDM printing flexible materials and slice their prints into a gcode file. The laboratory assistant will provide print parameters and best practices in 3D printing to students.

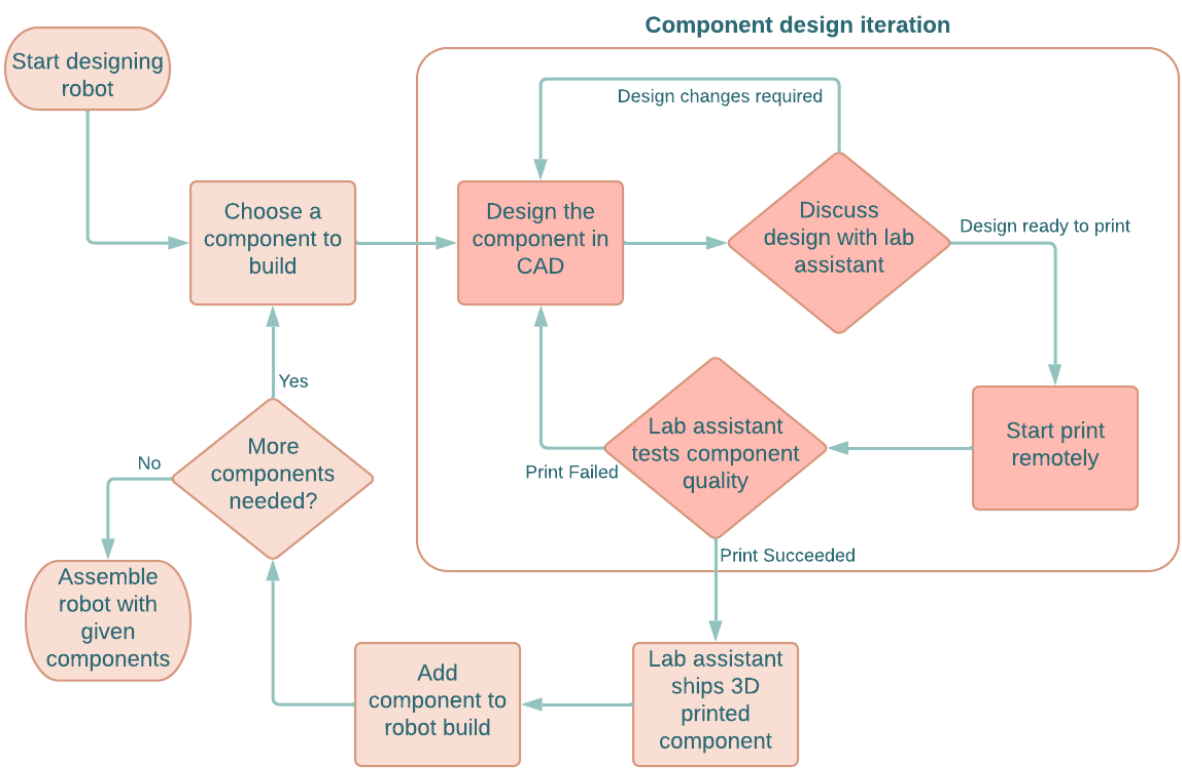

**Figure 4: A flowchart of using 3D printing for soft robotics for remote learning.** Students choose a component of their soft robot (i.e., sensor, actuator, or controller), and design and discuss their implementation with lab staff. Once the design is approved, it is sent to the printing station, and the print is tested together with the lab assistant. Successful prints are being shipped to students and once all components have been printed and shipped, the entire soft robot can be assembled by students at home.

B.  **Web-based interface for remote printing and monitoring**

The lab instructor will connect 3D printers, and one camera per printer, to a web-based interface using open-source software (e.g., Repetier or OctoPrint). Students will be granted access to the web interface to connect to the 3D printers remotely. After consultation with the lab assistant, students will slice their designs, taking into consideration the model of the 3D printer and printed material. Once students uploaded their prints to the web-based interface, they can monitor the progress of their print through the web camera feed.

### C. Inspection and delivery by teaching staff

Lab staff will remove the component from the printer, inspect it for print failures, potentially remove support materials, and label the component. Lab staff can evaluate the quality of the component by performing characterization tests as requested by the student. If a student is unsatisfied with the characteristics of the component, they will iterate on the design and print a new version of their component. When the component operates as intended, the lab staff will ship the print to the student using a service such as USPS, FedEx, or Sendle. It costs approximately $4 to $10 to send a package using these services.

### D. Assembly and testing at home

Students will receive their components through mail. They will inspect and test each component to understand the functionality of soft structures. Students can use this hands-on experience with a flexible material to design their next component. Once all components are printed and shipped, students will complete their assembly by interconnecting pneumatic channels with tubes or connectors supplied in the shipments. To actuate pneumatically driven soft robots, students can buy a low-cost pump for approximately $20 (12 V diaphragm pump, Amazon.com). The actuation pressure depends on the material of the filament and the structure of the soft device. Students can experiment with their assembled robot and observe the capabilities of soft robots.

## Discussion

### A. Comparison of the methodology to existing robotic kits

Many robotics kits exist for mobile robots that can be incorporated into teaching robotic courses. These robotic kits can be challenging to use in remote learning without an immediate availability of a lab. The modular design of these robots (e.g., Adafruit's 2 wheeled robot, Romi robot) is beneficial to students in-person when it is easy to receive a new component from a bulk supply in the lab; however, when remote, the failure of a single part requires students to purchase and ship a new component. Limitations in shipping and backlogs from vendors can hinder students' progress. Lastly, students will have trouble when debugging electronics without equipment such as oscilloscopes, multimeters, or function generators. Resources, such as The Soft Robotics Toolkit, exist to aid educators to create various designs such as grippers, cord-activated fingers, or pneumatic wrist braces, among others.

3D printed components that students will create with our methodology not only allow students to cultivate their creativity and ingenuity but will also be lower in cost than electronic components. Costs for a traditional mobile robot for education range from $40 to $400 [13]. A single PneuNet actuator requires approximately 15 grams of filament. When a high-quality filament such as Ninjaflex 85A from NinjaTek is used, the PneuNet will need $0.69 worth of material to print.

### B. How does this methodology apply to other engineering courses?

We envision extrapolation of our methodology to other engineering fields such as biomedical engineering, electrical engineering, and aerospace engineering using a remote 3D printing station. For example, the biomedical engineering discipline includes courses on soft bendable continuous robots to help doctors with surgery or inspection of body parts [14]. Recent work on conductive filaments allow students from electrical engineering to develop their own circuit diagrams and work on flexible electronic circuits using multi-material FDM printers [15]. SLA printers with ceramic resins can be post-processed in a kiln to produce small rocket parts, allowing aerospace students to experiment with nozzle characteristics and designs of rocket parts [16].

### C. Limitations of 3D printing soft robots

Although FDM printing with flexible materials makes our methodology scalable, FDM printing is not always the optimal solution for the fabrication of soft robots. We are limited by the filaments that are currently commercially available and their material properties. For example, Filaflex 60A Pro is the most elastic 3D printing filament currently on the market with a shore hardness of 60A (softness of car tire rubber). In comparison, additive curing elastomers (e.g., EcoFlex 00-10) can have a shore hardness as low as 00-10A (the same softness as gummy bears). Printing flexible materials also requires expertise. Some common problems that occur when working with flexible filaments are increased print times and entangled filament in the feed gears. Other printing techniques such as SLA, can produce higher quality prints of softer materials, however, they require time-consuming printing protocols including post-processing of components. We chose FDM printing due to its low cost, and availability and scalability for classrooms. Soft robots can be made from an integrated modular structure, in which our actuators can be 3D printed using flexible filaments and can be controlled via conventional control means.

### D. Limitations with lab logistics

Other limitations to this methodology come from the logistics of running a fabrication lab and from shipping. The number of 3D printers available in a laboratory affects the number of parts that can be produced simultaneously. A trained lab staff member must be around in case of failed prints; otherwise, students can also inform lab staff of print defects by watching their printer remotely. The lab staff will also have to remove completed prints and load filaments into the printers. However, we believe that careful scheduling of lab staff and prints will circumvent most of these issues.

The cost and time of shipping directly impact our methodology. Students in different parts of the country could receive their parts at different times and costs. Shipping to different locations within the United States costs about the same, but the cost dramatically increases with expedited shipping. The instructor can minimize shipping costs by consolidating components where possible. Students will mitigate the need for immediate shipping by working with the lab staff to characterize their components and iterate rather than shipping every print. In general, we believe this methodology will succeed when executed by schools within their locality, whereas experiences of this methodology can be exchanged at international conferences.

**Conclusion**

We introduced a remote learning methodology that utilizes printable, flexible robots to cultivate a hands-on learning experience, where robotics engineering students can design, fabricate, and assemble their robots. Students will learn fundamental skills such as design, fabrication, testing, and characterization of robotics components. We propose learning methods based on 3D printing, that is becoming an essential technology for customized robots. Working with printable flexible robots enables students to broaden their horizons from conventional robots. With an existing infrastructure of 3D printers, the cost of our methodology is at a minimum; it is solely the price of the flexible filament and the price of shipping. Students will gather experience with FDM printing of flexible filaments, iterative design, and hands-on experiments with their robots being off-campus. Our methodology of remote printing of flexible components gives access to robotics engineering education to students by decoupling the need for access to laboratory equipment.